\documentclass{ifacconf}

\usepackage{amssymb}
\usepackage{amsmath}
\usepackage{float}
\usepackage{mathtools}

\usepackage[font=footnotesize,labelfont=footnotesize]{subcaption}
\usepackage{comment}
\usepackage{multirow}
\usepackage{environ}
\usepackage{graphicx}

\usepackage{ algorithm, algorithmicx} 
\usepackage[noend]{algpseudocode}

\graphicspath{{img/}}

\theoremstyle{plain} 
\newtheorem{theorem}{Theorem}

\algnewcommand\algorithmicinput{\textbf{Input:}}
\algnewcommand\algorithmicoutput{\textbf{Output:}}
\algnewcommand\Input{\item[\algorithmicinput]}
\algnewcommand\Output{\item[\algorithmicoutput]}

\DeclareMathOperator*{\argmin}{arg\,min}

\usepackage{xcolor}

\newcommand{\remove}[1]{}

\usepackage{natbib}        
\begin{document}
\begin{frontmatter}

\title{Occlusion-Free Image-Based Visual Servoing using Probabilistic Control Barrier Certificates} 

\thanks[footnoteinfo]{This work was supported in part by the Faculty Research Grant award
at the University of North Carolina at Charlotte.}

\author[First]{Yanze Zhang}
\author[First]{Yupeng Yang}
\author[First]{Wenhao Luo} 

\address[First]{Department of Computer Science, University of North Carolina at Charlotte, Charlotte, NC 28223 USA\\ (e-mail: yzhang94,yyang52,wenhao.luo@uncc.edu).}

\begin{abstract} 
Image-based visual servoing (IBVS) is a widely-used approach in robotics that employs visual information to guide robots towards desired positions. However, occlusions in this approach can lead to  visual servoing failure and degrade the control performance due to the obstructed vision feature points that are essential for providing visual feedback. In this paper, we propose a Control Barrier Function (CBF) based controller that enables occlusion-free IBVS tasks by automatically adjusting the robot's configuration to keep the feature points in the field of view and away from obstacles. In particular, to account for measurement noise of the feature points, we develop the Probabilistic Control Barrier Certificates (PrCBC) using control barrier functions that encode the chance-constrained occlusion avoidance constraints under uncertainty into deterministic admissible control space for the robot, from which the resulting configuration of robot ensures that the feature points stay occlusion free from obstacles with a satisfying predefined probability. By integrating such constraints with a Model Predictive Control (MPC) framework, the sequence of optimized control inputs can be derived to achieve the primary IBVS task while enforcing the occlusion avoidance during robot movements. Simulation results are provided to validate the performance of our proposed method.
\end{abstract}

\end{frontmatter}

\section{Introduction}
Image-Based Visual Servoing (IBVS) is a key problem in robotics that involves using image data to control the robot's movement to a desired location. 
Specifically, this involves selecting specific feature points in images to generate a sequence of motions that move the robot in response to observations from a camera, ultimately reaching a goal configuration in the real world \citep{chaumette2016visual}. Many IBVS methods have been proposed and used in real-world applications such as the Visual Servoing Platform \citep{Marchand05b}, the Amazon Picking Challenge \citep{huang2018case}, and Surgery \citep{li2020accelerated}. However, these methods often assume that there are no obstacles obstructing the camera's field of view (FoV), allowing feature points to remain visible during servoing. When obstacles enter the camera's FoV, direct feedback from objects in the scene may be lost, making navigation more difficult.

To address the problem, some researchers focus on using environment knowledge to design the robot controller in case there is occlusion during visual servoing. One idea is using feature estimation method to recover control properties during occlusions, either by estimating point feature depth with nonlinear observers \citep{de2008feature}, or using a geometric approach to reconstruct a dynamic object characterized by point and line features \citep{fleurmond2016handling}. Another idea is to plan a camera trajectory that would altogether avoid the occlusion by other obstacles. \citep{kazemi2010path} presented an overview of the main visual servoing path-planning techniques which can be used to guarantee occlusion-free and collision-free trajectories, as well as to consider FoV limitations. However, these methods rely on accurate knowledge of the (partial) workspace and may require long computation time.

Other researchers used potential fields \citep{mezouar2002avoiding} or variable weighting Linear Quadratic control laws \citep{kermorgant2013dealing} to preserve visibility and avoid self-occlusions. Although these approaches are suitable for real-time implementation, they may exhibit local minima and possible unwanted oscillations. Moreover, the existing methods generally do not address the uncertainty in the environment, which could easily jeopardize the performance guarantee in presence of realistic factors such as measurement noises on the image.

On the other hand, Model Predictive Control (MPC) has been widely used in robotic systems to generate a sequence of controller by considering a finite time horizon optimization. It can deal with constraints, non-minimum phase processes and implement robust control even if the system dynamics is time-varying.
A large amount of work about MPC based visual servoing has been studied \citep{saragih2019visual,nicolis2018occlusion}.

Control Barrier Functions (CBFs) based methods have been used for safety critical applications such as automobile \citep{xu2017realizing} and human-robot interaction \citep{landi2019safety} with its provable theoretical guarantees that allow for a forward invariant desired set. CBFs have been used in visual servoing to keep the target in the FoV in \citep{zheng2019toward}. However, this method is not suitable for handling occlusion problems and does not account for possible measurement uncertainties.

In this paper, we present a control method that enables a chance-constrained occlusion-free guarantee for IBVS tasks under camera measurement uncertainty through the use of Probabilistic Control Barrier Certificates (PrCBC). The key idea of PrCBC, adopted in our IBVS problem, is to enforce the chance-constrained occlusion avoidance between the feature points and obstacle in the camera view with deterministic constraints over an existing robotic controller, so that the occlusion-free movements can be achieved with a satisfying probability.
We further integrate the PrCBC control constraints into a standard model predictive control (MPC) framework. 
Without loss of generality, we take the general case of a 6-DOF robot arm with eye-in-hand configuration as an example for the IBVS task, and provide simulation results on such platform to demonstrate the effectiveness of our proposed approach.
Our key insight in this paper is that the proposed PrCBC
can 
filter out the unsatisfying robot control action from the primary IBVS controller that may lead to occlusion with the obstacles, and leverage optimal controller from the integrated MPC framework. This insight leads us to the following contributions:

\begin{enumerate}
    \item We present a novel chance-constrained occlusion avoidance method for
    visual servoing tasks
    using Probabilistic Control Barrier Certificates (PrCBC) under camera measurement uncertainty, with theoretical analysis on the performance guarantee.
    \item We integrate PrCBC with Model Predictive Control (MPC) to provide a high-level planner with a minimally invasive control behavior to guarantee the occlusion avoidance for IBVS task, and validate through extensive simulation results.
\end{enumerate}

\section{Preliminaries}
\subsection{Feature Points Dynamics}
Consider a moving camera which is fixed on the robot end-effector with the eye-in-hand configuration and the camera is viewing objects in the workspace.
$\mathcal{F}_c$ is a right-handed orthogonal coordinate frame whose origin is at the principle point of the camera and $Z$ axis is collinear with the optical axis of the camera. Assuming $\mathbf{q}_i=[x_i,y_i]^{\rm T}\in\mathbb{R}^2$ is the pixel coordinate of a static point in the image plane,
then we can define its normalized image plane coordinate as
$\bar{\mathbf{q}}_i\in\mathbb{R}^2$ according to the camera projection model \citep{corke2011robotics}.
\begin{equation}
    \label{pinhole}
    \bar{\mathbf{q}}_i =\left[{\begin{array}{cc} \frac{X_i}{Z_i}, \frac{Y_i}{Z_i} \end{array} }   \right]^{\rm T}  = \left[ {\begin{array}{cc} {f}&0\\ 0&{f} \end{array}} \right]^{-1} (\mathbf{q}_i - \left[ {\begin{array}{c} {p_x}\\{p_y}  \end{array} }\right ])
\end{equation}
where $f,p_x,p_y \in \mathbb{R}$ are camera intrinsic parameters 
and $[X_i,Y_i,Z_i]^{\rm T}$ is the 3D coordinate
of the point 
in $\mathcal{F}_c$. 

The image vision feature can be constructed from the normalized image plane coordinates such that $\mathbf{p}_i=\bar{\mathbf{q}}_i$. Assume there are $m$ feature points extracted from the image and the state vectors of current and target vision feature points
are denoted as $\mathbf{s}(t)=\left[\mathbf{p}_1(t);\cdots;\mathbf{p}_m(t)\right] \in  \mathbb{R}^{2m}$ at time $t$ and $\mathbf{s}^*=\left[\mathbf{p}_1^*; \cdots ;\mathbf{p}_m^*\right]\in  \mathbb{R}^{2m}$ respectively, then the IBVS task aims to regulate the positional image feature points error vector $\mathbf{e}(t) = \mathbf{s}(t) - \mathbf{s}^*$ to zero through robot movements, which thus drives the robot to the desired position.

According to \citep{chaumette2006visual}, the dynamics of the $m$ feature points can be expressed as:
\begin{equation}
    \label{control}
    \dot{\mathbf{s}} = \mathbf{L}_s(Z) \mathbf{V}_c
\end{equation}
where $\mathbf{L}_s=\left[\mathbf{L}_{s_1} \cdots \mathbf{L}_{s_m}\right]^{\rm T}\in\mathbb{R}^{2m \times d}$ is the image interaction matrix and can be computed refer to \citep{chaumette2006visual}, $\mathbf{V}_c\in\mathbb{R}^d$ is a vector of the robot motion controller to express the 
camera translation and rotation velocity in the workspace and $Z = [Z_1 \cdots Z_m]^{\rm T}\in \mathbb{R}^m$ denotes the 
depths of $m$ feature points. 
In this paper, we assume the depth information $Z$ of the feature points has been acquired. Therefore in IBVS, (\ref{control}) represents the system dynamics of the $m$ feature points with $\mathbf{s}$ as the system state and $\mathbf{V}_c$ as the system controller. 

Since $\mathbf{s}^*$ is predefined and time independent, we have:
\begin{equation}
    \dot {\mathbf{e}} = \dot {\mathbf{s}} = \mathbf{L}_s \mathbf{V}_c \label{eq:erdyna}
\end{equation}
According to \citep{chaumette2006visual}, the unconstrained gradient-based IBVS controller driving $\mathbf{e}\to 0$ can be defined as:
\begin{equation}
    \mathbf{V}_c=-\alpha {\mathbf{L}_s}^+ \mathbf{e} \label{eq:controller}
\end{equation}
where $\alpha \in \mathbb{R}$ is a constant as predefined control gain, and ${{\mathbf{L}}_s}^+\in\mathbb{R}^{d \times 2m}$ is the pseudo-inverse of $\mathbf{L}_s$, which is given by ${{\mathbf{L}}_s}^+ = {(\mathbf{L}^{\rm T}_s\mathbf{L}_s)}^{-1}{\mathbf{L}_s}^{\rm T}$. To ensure the local asymptotic stability, 
we should have $\mathbf{L}^{\rm T}_s\mathbf{L}_s>0$, i.e. 
at least three 
feature points should be available in the FoV.

\subsection{Model Predictive Control Policy}
\label{sec:MPC}
To optimize the IBVS controller over a finite
time horizon, the MPC policy is used to construct a $N$ step time horizon planner rendering a sequence of candidate control actions. 
Consider the problem of IBVS that seeks to regulate the current feature points to the target feature points 
through robot movements. According to 
the error dynamics (\ref{eq:erdyna}), the discrete-time control system can be described by:
\begin{equation}
    \mathbf{e}(t+1) =\mathbf{I}\mathbf{e}(t)+\mathbf{L}_s\mathbf{V}_c(t)= g(\mathbf{e}(t),\mathbf{V}_c(t))
\end{equation}
where $\mathbf{I}\in\mathbb{R}^{2m \times 2m}$ is an identity matrix and $\mathbf{e}(t)\in\mathbb{R}^{2m}$ represents the state of the image feature error of $m$ feature points at time step 
$t$.
The system state is $\mathbf{s}(t)$ with $\mathbf{V}_c(t)\in\mathcal{U}\subset \mathbb{R}^d$ as the control input and $g$ is locally Lipschitz.

Therefore, the finite-time optimal control problem can be solved at time step $t$ using the following policy $\pi$:
\begin{footnotesize}
\begin{align}
    &
    \min_{\mathbf{V}_c(t:t+N-1)}\{ \sum_{k=0}^{N-1}(\mathbf{e}({t+k|t})^{\rm T}\mathbf{Q} \mathbf{e}({t+k|t})\\
    &\quad +\mathbf{V}_c({t+k|t})^{\rm T}\mathbf{R}\mathbf{V}_c({t+k|t})) +\mathbf{e}({t+N|t})^{\rm T}\mathbf{F} \mathbf{e}({t+N|t})\} \notag \label{costfunc}\\
    \text{s.t.}\;& \mathbf{e}(t+k+1|t+k)=g(\mathbf{e}(t+k|t),\mathbf{V}_c(t+k|t))   
    \\ 
    &\mathbf{e}(t+k|t)\in\mathcal{X},\mathbf{V}_c(t+k|t)\in\mathcal{U},k=0,\ldots,N-1\\
    &\mathbf{e}(t|t)=\mathbf{e}(t)
\end{align}
\end{footnotesize}
where $N$ is the prediction time horizon, and $\mathbf{Q}\in\mathbb{R}^{2m \times 2m}$, $\mathbf{R} \in \mathbb{R}^{d \times d}$ and $\mathbf{F}\in\mathbb{R}^{2m \times 2m}$ are weighting matrices which represent a trade-off between the small magnitude of the control input (with larger value of $\mathbf{R}$) and fast response (with larger value of $\mathbf{Q}$ and $\mathbf{F}$). $\mathbf{e}(t+k|t)$ denotes the error vector at time step $t+k$ predicted at time step $t$, which is obtained from the current image feature error $\mathbf{e}(t)$ and the control input of $\mathbf{V}_c({t:t+N-1})$. 
Finally, the optimized control sequence can be obtained as $\mathbf{V}_c^{t:t+N-1}$.

\subsection{Obstacle Model and Occlusion Avoidance} 
Without loss of generality, 
Consider an obstacle $O$ which is moving in the workspace and may occlude the feature points in the camera view. 
We model the obstacle as a rigid sphere with the radius $R$ in the workspace. 
Similar to the feature points, the normalized image plane coordinates of the obstacle center and radius of the obstacle
can be defined as
$\mathbf{s}_\mathrm{o}(t)=[\frac{X_\mathrm{o}}{Z_\mathrm{o}(t)},\frac{Y_\mathrm{o}}{Z_\mathrm{o}(t)}]^{\rm T} \in \mathbb{R}^2$
and $R_n(t) = \frac{R}{Z_\mathrm{o}(t)}$ respectively. To simplify the notation,
we will use $\mathbf{s}_o$ and $R_n$ to denote the real-time state of
the obstacle center and obstacle radius in the normalized image plane.

In this way, for any pair-wise occlusion avoidance between the feature point $i$
and the obstacle $O$ in the normalized image plane,
the occlusion-free condition and state set can be defined as follows:
\begin{equation}
    h^c_{i,\mathrm{o}}(\mathbf{s},\mathbf{s}_\mathrm{o})={\left \| \mathbf{s}_i-\mathbf{s}_\mathrm{o} \right \|}^2-R_n^2,\forall i \label{control_barrier_function}
\end{equation}
\begin{equation}
    \label{occlusion-free}
    \mathcal{H}^c_{i,\mathrm{o}}= \left\{\mathbf{s}_i,\mathbf{s}_\mathrm{o}\in\mathbb{R}^{2}| h^c_{i,\mathrm{o}}(\mathbf{s},\mathbf{s}_\mathrm{o})\ge0  ,\forall i\right\}
\end{equation}
Then for all feature points, the desired set for occlusion-free states is thus defined as:
\begin{equation}
\label{eq:occlusion-free-all}
    \mathcal{H}^c = \bigcap_{\{\forall i\}} \mathcal{H}^{c}_{i,\mathrm{o}}
\end{equation}

\subsection{Occlusion-free Constraints using CBF}
Control Barrier Functions (CBF) \citep{ames2019control}  have been widely applied to generate control constraints that render a set forward invariant, i.e. if the system state starts inside a set, it will never leave this set under the satisfying controller. The main idea of CBF is summarized as the following Lemma.
\begin{lem}
\label{Lemma1}[summarized from \cite{ames2019control}]
Given a dynamical system affine in control and a desired set $\mathcal{H}$ as the 0-superlevel set of a continuous differentiable function $h(\mathbf{x}) : \mathcal{X} \rightarrow \mathbb{R}$, the function $h$ is a control barrier function if there exists an extended class-$\mathcal{K}$ function $\kappa(\cdot)$ such that $\sup_{\mathbf{u}\in\mathcal{U}}[\dot{h}(\mathbf{x},\mathbf{u})+\kappa(h(\mathbf{x}))]\geq 0$ for all $\mathbf{x}$. The admissible control space $\mathcal{B}(\mathbf{x})$ for Lipschitz continuous controller $\mathbf{u}$ rendering $\mathcal{H}$ forward invariant (i.e. keeping the system state $\mathbf{x}$ staying in $\mathcal{H}$ over time) thus becomes:
\begin{equation}\label{eq:cbc_lemma}
    \mathcal{B}(\mathbf{x}) = \{ \mathbf{u}\in \mathcal{U} | \dot{h}(\mathbf{x},\mathbf{u}) + \kappa(h(\mathbf{x}))\geq 0 \}
\end{equation}
\end{lem}
Based on the occlusion avoidance condition ~(\ref{control_barrier_function})-(\ref{occlusion-free}) and the Lemma~\ref{Lemma1}, if the feature points are not occluded by the obstacle initially, then the admissible control space for the robot to keep the feature points free from occlusion
can be represented by the following control constraints over $\mathbf{V}_c$:
\begin{footnotesize}
\begin{align}
\label{eq:occlusion-free CBC}
    \mathcal{B}(\mathbf{s},\mathbf{s}_\mathrm{o}) = \{ \mathbf{V}_c\in \mathcal{U} | \dot {h}^c_{i,\mathrm{o}}(\mathbf{s},\mathbf{s}_\mathrm{o}, \mathbf{V}_c) + \gamma(h^c_{i,\mathrm{o}}(\mathbf{s},\mathbf{s}_\mathrm{o}))\geq 0, \forall i\}
\end{align}
\end{footnotesize}
where $ \mathcal{B}(\mathbf{s},\mathbf{s}_\mathrm{o})$ defines the Control Barrier Certificates (CBC) for the feature point-obstacle occlusion avoidance, and
$\gamma$ is a user-defined parameter in the particular choice of $\kappa(h(\mathbf{x})) = \gamma(h(\mathbf{x}))$ as in \citep{luo2020behavior}.
It will render the occlusion-free set $\mathcal{H}^c$ forward invariant, i.e. as long as we can guarantee the control input $\mathbf{V}_c$ lies in the set $\mathcal{B}(\mathbf{s},\mathbf{s}_\mathrm{o})$, the feature points will not be occluded by the obstacle at all times.

\subsection{Chance Constraints for Measurement Uncertainty}
Although (\ref{eq:occlusion-free CBC}) indicates an explicit condition for occlusion avoidance with the perfect knowledge of the actual position $\mathbf{s}$, the presence of measurement uncertainty on $\mathbf{s}$ makes it challenging to enforce (\ref{eq:occlusion-free CBC}), or even impossible when the measurement noise is unbounded, e.g. Gaussian noise. In this paper, we consider the realistic situation where the pixel coordinates of the extracted feature points acquired by the camera have Gaussian distribution noise that are formulated as follows:
\begin{equation}\footnotesize
    \hat{\mathbf{s}}_i = \mathbf{s}_i + \mathbf{\mathbf{w}_i}, \mathbf{\mathbf{w}_i} \sim N(0,\Sigma_i),\;\hat{\mathbf{s}}_\mathrm{o} = \mathbf{s}_\mathrm{o} + \mathbf{\mathbf{w}_\mathrm{o}}, \mathbf{\mathbf{w}_\mathrm{o}} \sim N(0,\Sigma_\mathrm{o})     
\end{equation}
where ${\mathbf{w}_i,\mathbf{w}_\mathrm{o}}\in\mathbb{R}^{2}$ are the Gaussian measurement noises and can be considered as independent random variables with zero mean and $\Sigma_i,\Sigma_\mathrm{o}$ as the variances. With that, for the rest of the paper we assume only the noisy positions $\hat{\mathbf{s}}_i$ and $\hat{\mathbf{s}}_\mathrm{o}$ of extracted feature points are available when the uncertainty is considered.

Then the occlusion avoidance condition in (\ref{control_barrier_function})-(\ref{occlusion-free}) can be considered in a chance-constrained setting. Formally, given the user-defined satisfying probability threshold of the occlusion avoidance as
$\sigma \in (0,1)$, we have:
\begin{equation}
    {\rm{Pr}}(\mathbf{s},\mathbf{s}_\mathrm{o}\in\mathcal{H}^c)\ge\sigma \label{eq:chance}
\end{equation}
where \rm{Pr}(.) indicates the probability of an event. Note that when the $\sigma$ is approaching 1, it will lead to a more conservative controller to maintain the probabilistic occlusion-free set.
In Section~\ref{sec:PrCBC}, we will discuss how to transfer the chance constraints on the feature points states into a deterministic admissible control space $\mathcal{S}^{\sigma}(\mathbf{\hat{s}},\mathbf{\hat{s}}_\mathrm{o})\subset \mathcal{U}$ w.r.t. $\mathbf{V}_c$, so that the occlusion-free performance could be guaranteed with satisfying probability as implied in (\ref{eq:chance}).

\subsection{Problem Statement}
\label{sec:problem}
To achieve the occlusion-free IBVS, we first use the MPC as a planner to generate the unconstrained control sequence $\mathbf{V}_c^{t:t+N-1}$ at each time step $t$. 
To address the occlusion problem brought by the obstacle, we discuss two situations to define the optimized problem.

\textbf{Without Noise}: First, we assume that the camera can get the accurate pixel coordinates from the image, i.e. we can extract the feature points precisely. In this way, we assume the feature points are not occluded by the obstacle at the initial location.
Then we can formally define this occlusion-free constraint with the following step-wise Quadratic-Program (QP) at each time step $t$:
\begin{align}
 &\mathbf{V}_c^* = \argmin_{\mathbf{V}} ||\mathbf{V}-\mathbf{{V}}_c^{mpc}||^2 \label{eq:withoutnoise}\\
 &\text{s.t.} \quad \mathbf{V}\in \mathcal{B}(\mathbf{s},\mathbf{s}_\mathrm{o}) \quad  ||\mathbf{V}|| \leq {V}_{\mathrm{max}}
\end{align}
where $\mathbf{V}_c^{mpc} \in \mathbb{R}^d$ is the first step of the control sequence $\mathbf{V}_c^{t:t+N-1}$ generated from MPC at time $t$, and ${V}_{\mathrm{max}}$ is the maximum velocity. 
Hence the resulting $\mathbf{V}_c^* \in \mathbb{R}^d$ is the step-wise optimized controller to be executed at each $t$ that renders the occlusion-free set $\mathcal{H}^c$ in (\ref{eq:occlusion-free-all}) forward invariant. 

\textbf{With Noise}: In presence of camera measurement noise, then only the inaccurate vision feature information of the normalized image plane $\hat{\mathbf{s}}_i$ and $\hat{\mathbf{s}}_\mathrm{o}$ can be obtained for IBVS task. In this case, the chance-constrained occlusion-free problem can be formally defined as:
\begin{align}
 &\mathbf{V}_c^* = \argmin_{\mathbf{V}} ||\mathbf{V}-\mathbf{{V}}_c^{mpc}||^2 \label{eq:withnoise}\\
 &\text{s.t.} \quad \mathbf{V}\in \mathcal{S}^{\sigma}(\mathbf{\hat{s}},\mathbf{\hat{s}}_\mathrm{o}), \quad ||\mathbf{V}|| \leq {V}_{\mathrm{max}}
\end{align}

\section{Method}
\begin{figure}
\captionsetup{skip=0pt}
  \centering
\includegraphics[width=2.5in]{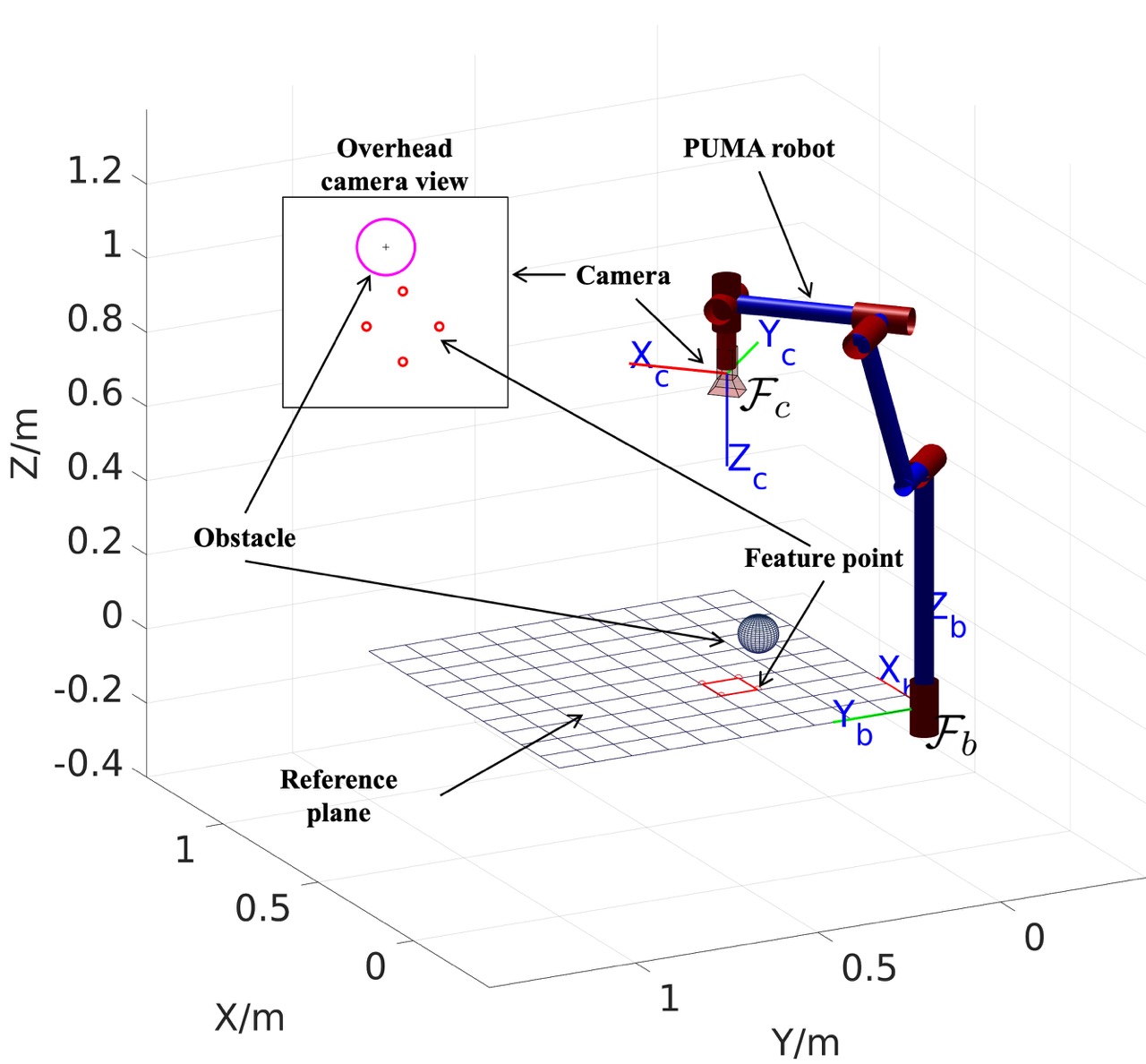}
\caption{The IBVS scenario with a moving obstacle.}
\label{fig:scenario}
\end{figure}

We consider the IBVS task scenario as shown in Fig.~\ref{fig:scenario}, where the camera is attached to a 6-DOF \emph{PUMA} robot end-effector and four feature points are set (i.e. $m=4$) as the extracted image feature for guiding the IBVS task as commonly assumed e.g. \citep{chaumette2006visual}. 
There is one moving obstacle with radius $R \in \mathbb{R}$ in the workspace which may occlude the feature points during the execution of the primary IBVS task. Therefore, the objective of the IBVS task for the robot to achieve a desired configuration is to move the \emph{PUMA} robotic arm 
in a way that the feature points positions in the camera view eventually converge 
to the desired positions without occlusion from the obstacle at all times.

With that, we have the feature points state as  $\mathbf{s}(t)=[\mathbf{p}_1(t);\mathbf{p}_2(t); \mathbf{p}_3(t);\mathbf{p}_4(t)] \in \mathbb{R}^8$ and the corresponding control input 
$\mathbf{V}_c = \left[v_x,v_y,v_z,\omega_x,\omega_y,\omega_z\right]^{\rm T}$ that expresses the 6-DOF motion controller of the camera. 
$[v_x,v_y,v_z]^{\rm T}$ and $[\omega_x,\omega_y,\omega_z]^{\rm T}$ are the vector of the linear and angular velocities.
$\mathbf{L}_s = \left[{\mathbf{L}_{s1}}; {\mathbf{L}_{s2}}; {\mathbf{L}_{s3}}; {\mathbf{L}_{s4}}\right]\in\mathbb{R}^{8 \times 6}$ is the interaction matrix which is detailed later in (\ref{eq:interaction}). Similar to $\mathbf{s}$, we can acquire the state of the obstacle center as $\mathbf{s}_\mathrm{o}(t) = \mathbf{p}_\mathrm{o}(t)$.
According to \citep{chaumette2006visual}, we can calculate interaction matrix of $i$-th feature point:
\begin{equation}
\label{eq:interaction}
{\mathbf{L}_{{s_i}}} = \left[ {\begin{array}{cccccc} {\frac{{ - 1}}{{{Z_i}}}}&0&{\frac{{{p_{i,1}}}}{{{Z_i}}}}&{{p_{i,1}}{p_{i,2}}}&{ - \left( {1 + p_{i,1}^2} \right)}&{{p_{i,2}}} \\ 0&{\frac{{ - 1}}{{{Z_i}}}}&{\frac{{{p_{i,2}}}}{{{Z_i}}}}&{1 + p_{i,2}^2}&{ - {p_{i,1}}{p_{i,2}}}&{ - {p_{i,1}}} \end{array}} \right] 
\end{equation}
where $[p_{i,1},p_{i,2}]^{\rm T}=\mathbf{p}_i$ and $Z_i$ is the depth of $i$-th feature point in $\mathcal{F}_c$. 
Similarly, the interaction matrix of the obstacle center and obstacle radius, $\mathbf{L}_\mathrm{o}$ and $\mathbf{L}_\mathrm{or}$ can be derived as:
\begin{equation}
{\mathbf{L}_\mathrm{o}} = \left[ {\begin{array}{cccccc} {\frac{{ - 1}}{{{Z_\mathrm{o}}}}}&0&{\frac{{{p_{\mathrm{o},1}}}}{{{Z_\mathrm{o}}}}}&{{p_{\mathrm{o},1}}{p_{\mathrm{o},2}}}&{ - \left( {1 + p_{\mathrm{o},1}^2} \right)}&{{p_{\mathrm{o},2}}} \\ 0&{\frac{{ - 1}}{{{Z_\mathrm{o}}}}}&{\frac{{{p_{\mathrm{o},2}}}}{{{Z_i}}}}&{1 + p_{\mathrm{o},2}^2}&{ - {p_{\mathrm{o},1}}{p_{\mathrm{o},2}}}&{ - {p_{\mathrm{o},1}}} \end{array}} \right]
\end{equation}
\begin{equation}
{\mathbf{L}_\mathrm{or}} = \left[ {\begin{array}{cccccc} {0}&0&{\frac{R}{Z_\mathrm{o}^2}}&{\frac{{R}{p_{\mathrm{o},2}}}{Z_\mathrm{o}}}&{ - \frac{{R}{p_{\mathrm{o},1}}}{{Z_\mathrm{o}}} }&{0} \end{array}} \right]
\end{equation}
where $\mathbf{p}_\mathrm{o} = [p_{\mathrm{o},1},p_{\mathrm{o},2}]^{\rm T}$ and $Z_\mathrm{o}$ is obstacle depth in $\mathcal{F}_c$.

Intuitively, the distance will be used to derive the analytical form of the occlusion-free set. According to the different situations discussed in Section \ref{sec:problem}, the CBC and PrSBC for occlusion-free performance
are detailed as follows.

\subsection{Control Barrier Certificates for Occlusion Avoidance}
First, we consider the condition where the camera can acquire the accurate pixel coordinate from the world space when there is no measurement noise. 
Given the occlusion free condition in  (\ref{control_barrier_function})- (\ref{occlusion-free}) and the form of CBC in (\ref{eq:occlusion-free CBC}), we now formally define the CBC for the occlusion avoidance condition as follows:
\begin{theorem}\label{lemma2}
Given a desired occlusion-free set $\mathcal{H}^c$ in (\ref{occlusion-free}) with function $h^c_{i,\mathrm{o}}(\mathbf{s},\mathbf{s}_\mathrm{o})$ in (\ref{control_barrier_function}), the admissible control space for Lipschitz continuous controller defined below renders $\mathcal{H}^c$ forward invariant, i.e., keeping feature points away from the obstacle in the image plane over time.
\begin{footnotesize}
\begin{align}
    &\mathcal{B}(\mathbf{s},\mathbf{s}_\mathrm{o}) = \{ \mathbf{V}_c\in \mathcal{U} | \dot {h}^c_{i,\mathrm{o}}(\mathbf{s},\mathbf{s}_\mathrm{o},\mathbf{V}_c) + \gamma(h^c_{i,\mathrm{o}}(\mathbf{s},\mathbf{s}_\mathrm{o}))\geq 0, \forall i
    \}\\
    &\dot {h}^c_{i,\mathrm{o}}(\mathbf{s}_i,\mathbf{s}_\mathrm{o},\mathbf{V}_c) = 2(\mathbf{s}_{i}-\mathbf{s}_\mathrm{o})^{\rm T}(\mathbf{L}_{si}-\mathbf{L}_\mathrm{o})\mathbf{V}_c - 2 R_n\mathbf{L}_\mathrm{or}\mathbf{V}_c \label{eq:differntial}
\end{align}
\end{footnotesize}
\end{theorem}

\begin{pf}
We demonstrate that our proposed control barrier function $h^c_{i,\mathrm{o}}(\mathbf{s}_i,\mathbf{s}_\mathrm{o})$ in (\ref{control_barrier_function}) is a valid control barrier function.
As summarized in \citep{capelli2020connectivity}, from Lemma~\ref{Lemma1} the condition for a function $h(\mathbf{x})$ to be a valid CBF should satisfy the following three conditions: (a) $h(\mathbf{x})$ is continuously differentiable, (b) the first-order time derivative of $h(\mathbf{x})$ depends explicitly on the control input $\mathbf{u}$ (i.e. $h(\mathbf{x})$ is of relative degree one), and (c) it is possible to find an extended class-$\mathcal{K}$ function $\kappa(\cdot)$ such that {\footnotesize$\sup_{\mathbf{u}\in\mathcal{U}} \{\dot{h}(\mathbf{x},\mathbf{u})+\kappa(h(\mathbf{x}))\}\geq 0$} for all $\mathbf{x}$.

Hence, consider our proposed candidate CBF $h^c_{i,\mathrm{o}}(\mathbf{s}_i,\mathbf{s}_\mathrm{o})$ in (\ref{control_barrier_function}), and according to the differential function in (\ref{eq:differntial}), it is straightforward that the first order derivative of (\ref{control_barrier_function}) in the form of (\ref{eq:differntial}) depends explicitly on the control input $\mathbf{V}_c$. Thus, the function ${h}^c_{i,\mathrm{o}}(\mathbf{s}_i,\mathbf{s}_\mathrm{o})$ in (\ref{control_barrier_function}) is (a) continuously differential and (b) is of relative degree one. 

For condition (c) {\footnotesize $\sup_{\mathbf{V}_c\in\mathcal{U}}[\dot {h}^c_{i,\mathrm{o}}(\mathbf{s},\mathbf{s}_\mathrm{o},\mathbf{V}_c) + \gamma(h^c_{i,\mathrm{o}}(\mathbf{s},\mathbf{s}_\mathrm{o}))]\geq 0$}, we need to prove 
that 
the following inequality has at least one solution.
\begin{equation}
\label{pf:CBC}
    \dot {h}^c_{i,\mathrm{o}}(\mathbf{s},\mathbf{s}_\mathrm{o},\mathbf{V}_c) + \gamma(h^c_{i,\mathrm{o}}(\mathbf{s},\mathbf{s}_\mathrm{o})) \ge 0, \forall i \in \{1,2,3,4\}
\end{equation}
Given the form of $\dot {h}^c_{i,\mathrm{o}}(\mathbf{s},\mathbf{s}_\mathrm{o},\mathbf{V}_c)$ in (\ref{eq:differntial}), then (\ref{pf:CBC}) can be re-written as:
\begin{equation}
\label{eq:analytics}
   \mathbf{M} \mathbf{V}_c \ge n
\end{equation}
where {\footnotesize $\mathbf{M} = 2((\mathbf{s}-[\mathbf{s}_\mathrm{o};\mathbf{s}_\mathrm{o};\mathbf{s}_\mathrm{o};\mathbf{s}_\mathrm{o}])^{\rm T}(\mathbf{L}_{s}-[\mathbf{L}_\mathrm{o};\mathbf{L}_\mathrm{o};\mathbf{L}_\mathrm{o};\mathbf{L}_\mathrm{o}]) - R_n\mathbf{L}_\mathrm{or}) \in\mathbb{R}^{1\times6}$}, and {\footnotesize$n = \gamma (R_n^2 - {\left \| \mathbf{s}-[\mathbf{s}_\mathrm{o} ;\mathbf{s}_\mathrm{o};\mathbf{s}_\mathrm{o} ;\mathbf{s}_\mathrm{o} ]\right \|}^2) \in \mathbb{R}$}.
Given a specific state of $\mathbf{s}$ and $\mathbf{s}_o$, $n$ will be a constant. It is also straightforward that the six terms of {\footnotesize $\mathbf{M}$} can not be zero at the same time. Therefore, if we consider an unbounded control input {\footnotesize $\mathbf{V}_c \in \mathcal{U}\in \mathbb{R}^6$}, we can always find a solution such that {\footnotesize $\mathbf{M} \mathbf{V}_c\geq n$}. 

In the case that {\footnotesize $\mathbf{V}_c \in \mathcal{U}\in \mathbb{R}^6$} is bounded, several existing approaches could be employed to enforce the feasibility of the condition in (\ref{eq:analytics}). For example, in \citep{lyu2021probabilistic}, we provided an optimization solution for parameter $\gamma$ over time to guarantee that the admissible control set is always not empty when a feasible solution does exist. Besides, the authors in \citep{xiao2022sufficient} provided a novel method to find sufficient conditions, which 
are captured by a single constraint and enforced by an additional CBF, to guarantee the feasibility of the original CBF control constraint.
Readers are referred to \citep{xiao2022sufficient} for further details. With that, we conclude the proof. \hfill$\square$
\end{pf}

\subsection{Probabilistic Control Barrier Certificates (PrCBC) for Occlusion Avoidance} \label{sec:PrCBC}
In presence of uncertainty, similar to \citep{luo2020multi}, we have the sufficient condition of (\ref{eq:chance}) as
\begin{equation}\footnotesize
    \textbf{Pr}(\mathbf{V}_c \in \mathcal{B}(\mathbf{s}_i,\mathbf{s}_\mathrm{o})) \geq \sigma \Rightarrow \textbf{Pr}(\mathbf{s}_i \in \mathcal{H}^c_{i,\mathrm{o}}) \geq \sigma, \forall i\in\{1,2,3,4\}
\end{equation}
Following \citep{luo2020multi}, we define our PrCBC
for the chance-constrained occlusion free condition as follows:

\noindent
\textbf{Probabilistic Control Barrier Certificates (PrCBC):} Given a confidence level $\sigma \in (0,1)$, the admissible control space $\mathcal{S}^{\sigma}(\mathbf{\hat{s}},\mathbf{\hat{s}}_\mathrm{o})$ determined as below enforces the chance-constrained condition in (\ref{eq:chance}) at all times.
\begin{align}
\label{PrCBC}
    &\mathcal{S}^{\sigma}(\mathbf{\hat{s}},\mathbf{\hat{s}}_\mathrm{o})=\{ \mathbf{V}_c\in \mathcal{U} | \mathbf{V}_c^{\rm T} \mathbf{A}^\sigma_{i,\mathrm{o}}\mathbf{V}_c + \mathbf{b}^\sigma_{i,\mathrm{o}}\mathbf{V}_c + c_{i,\mathrm{o}}\leq 0, \notag\\
    &\mathbf{A}^\sigma_{i,\mathrm{o}} \in \mathbb{R}^{6 \times 6}, \mathbf{b}_{i,\mathrm{o}}^{\sigma}\in\mathbb{R}^{1 \times 6}, c_{i,\mathrm{o}} \in \mathbb{R}, \forall i\in\{1,2,3,4\} \}
\end{align}
The analytical form of $\mathbf{A}^\sigma_{i,\mathrm{o}} \in \mathbb{R}^{6 \times 6}, \mathbf{b}_{i,\mathrm{o}}^{\sigma}\in\mathbb{R}^{1 \times 6}, c_{i,\mathrm{o}} \in \mathbb{R}$ will be given in the latter part of (\ref{PrCBC2}).

\noindent
\textbf{Computation of PrCBC:}
Given the confidence level $\sigma \in (0,1)$, the chance constraints of {\footnotesize $\textbf{Pr}(\mathbf{V}_c \in \mathcal{B}(\mathbf{s}_i,\mathbf{s}_\mathrm{o})) \geq \sigma$} can be transformed into a deterministic quadratic constraint over the controller {\footnotesize $\mathbf{V}_c$} in the form of (\ref{PrCBC}). We denote  $e=\Phi^{-1}(\sigma)$ as one solution that makes the joint cumulative distribution function (CDF) of random variable of $\hat{\mathbf{s}}_i - \hat{\mathbf{s}}_\mathrm{o}$ equal to $\sigma$, where $e$ represent the side length of the square that covers the cumulative distribution probability with $\sigma$. Then, the deterministic constraints below become the sufficient condition for {\footnotesize$\textbf{Pr}(\mathbf{V}_c \in \mathcal{B}(\mathbf{s}_i,\mathbf{s}_\mathrm{o})) \geq \sigma$}:
\begin{footnotesize}
\begin{align}
    || \Delta \mathbf{s} ||^2 + 2\cdot \frac{\Delta \mathbf{s}^{\rm T}(\Delta \mathbf{L}-4Rr\mathbf{L}_r)\mathbf{V}_c}{\gamma} &\geq 2 R_n^2 + \frac{ \mathbf{V}_c^{\rm T} \Delta \mathbf{L}^{\rm T} \Delta \mathbf{L}  \mathbf{V}_c}{\gamma^2} \notag \\
    &+ 4e^2, \forall i \in \{1,2,3,4\}
\end{align}
\end{footnotesize}
where {\footnotesize$\Delta \mathbf{s} = \hat{\mathbf{s}}_i - \hat{\mathbf{s}}_\mathrm{o}$} and {\footnotesize $\Delta \mathbf{L} = \mathbf{L}_{s_i} - \mathbf{L}_\mathrm{o}$}. Therefore, we formally construct the PrCBC as the following deterministic quadratic constraints:
\begin{footnotesize}
\begin{align}
    \label{PrCBC2}
    \mathcal{S}^{\sigma}(\mathbf{\hat{s}},\mathbf{\hat{s}}_\mathrm{o})&=\{ \mathbf{V}_c\in \mathcal{U} | \frac{ \mathbf{V}_c^{\rm T} \Delta \mathbf{L}^{\rm T} \Delta \mathbf{L} \mathbf{V}_c}{\gamma^2} - 2\frac{\Delta \mathbf{s}^{\rm T}(\Delta \mathbf{L}-4R_n\mathbf{L}_r)\mathbf{V}_c}{\gamma} \notag \\&+ 2 R_n^2+ 4e^2-|| \Delta \mathbf{s} ||^2_2\leq 0, \forall i\in\{1,2,3,4\} \}
\end{align}
\end{footnotesize}

With that, we have {\footnotesize $\mathbf{A}^\sigma_{i,\mathrm{o}} = \frac{\Delta \mathbf{L}^{\rm T} \Delta \mathbf{L}}{\gamma^2}$, $\mathbf{b}_{i,\mathrm{o}} = \frac{-2\Delta \mathbf{s}^{\rm T}(\Delta \mathbf{L}-4R_n\mathbf{L}_r)}{\gamma}$} and {\footnotesize $c_{i,\mathrm{o}} =2 R_n^2+ 4e^2-|| \Delta \mathbf{s} ||^2_2$}.
Due to the size limitation of the physical workspace in the visual servoing scenario, the quadratic control constraint may not be feasible, e.g. when the moving obstacle is very large, it is impossible to avoid occlusion. In this extreme case, one alternative solution is to hold the robot static and to wait until the obstacle no longer occlude the feature points.

\subsection{MPC Planner with Occlusion Avoidance}
To integrate the advantages of the Model Predictive Control policy for high-level planning and CBC/PrCBC to enforce the occlusion-free condition, we combine the procedure of the MPC and proposed CBC/PrCBC to generate an optimized control sequence. Specifically, after executing the CBC/PrCBC to minimally modify the control $\mathbf{V}_c^{mpc}(t)$ to $\mathbf{V}_c^*(t)$ at time step $t$, the process is repeatedly implemented at the time step $t+1$: using policy $\pi$ in Section \ref{sec:MPC} to generate the control sequence $\mathbf{V}_c^{t+1:t+N|t+1}$, obtain the optimized control $\mathbf{V}_c^{mpc}(t+1)$, and calculate the occlusion-free control $\mathbf{V}_c^*(t+1)$ to execute next.

\section{Experiment Results and Discussion}
\begin{figure*}[!htbp]
\captionsetup{skip=0pt}
  \centering
  \begin{subfigure}{0.3\textwidth}
\includegraphics[width=\textwidth]{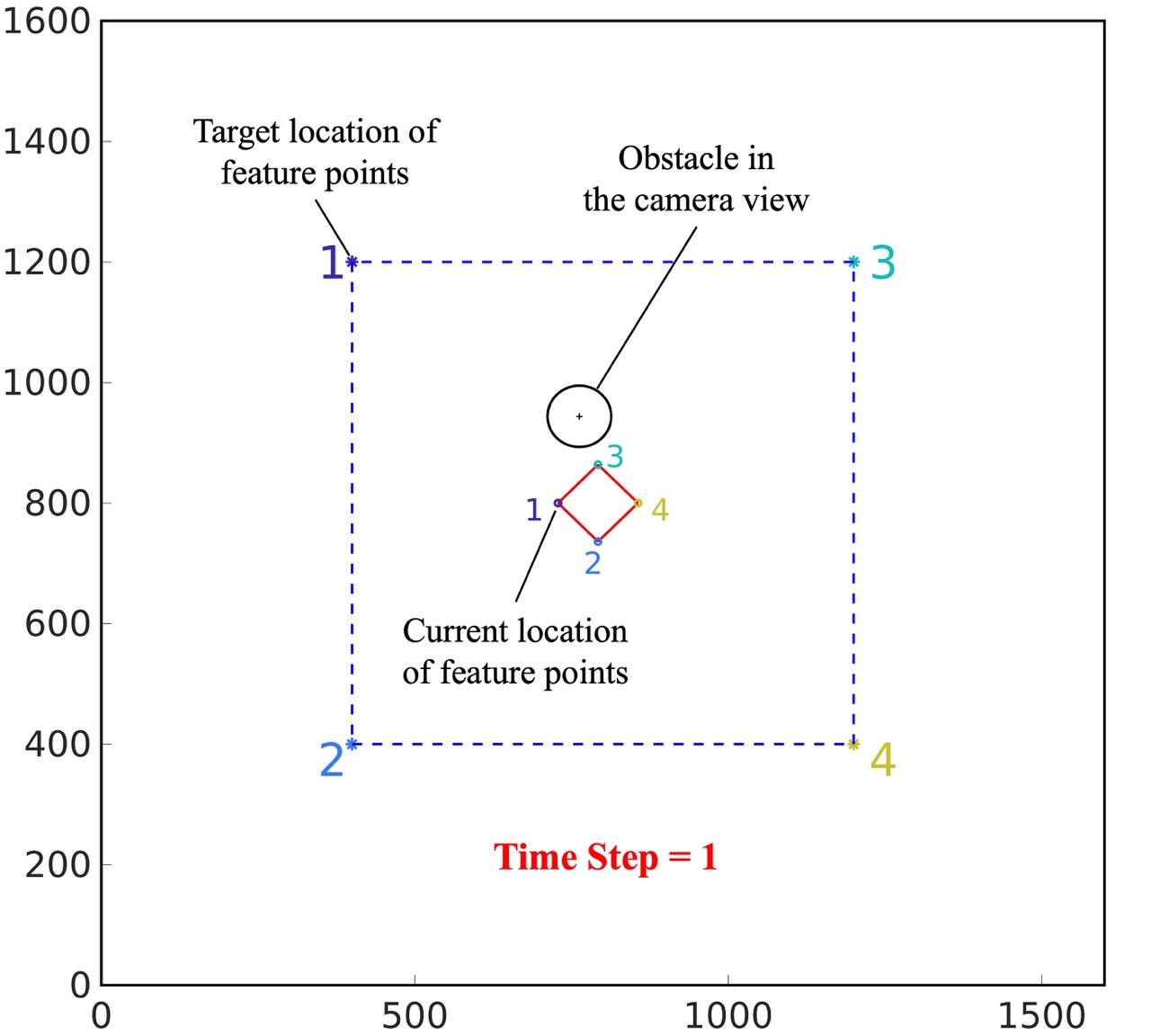}
    \caption{Initial time step for all experiments}
    \label{fig2:init}
  \end{subfigure}
  \begin{subfigure}{0.3\textwidth}
\includegraphics[width=\textwidth]{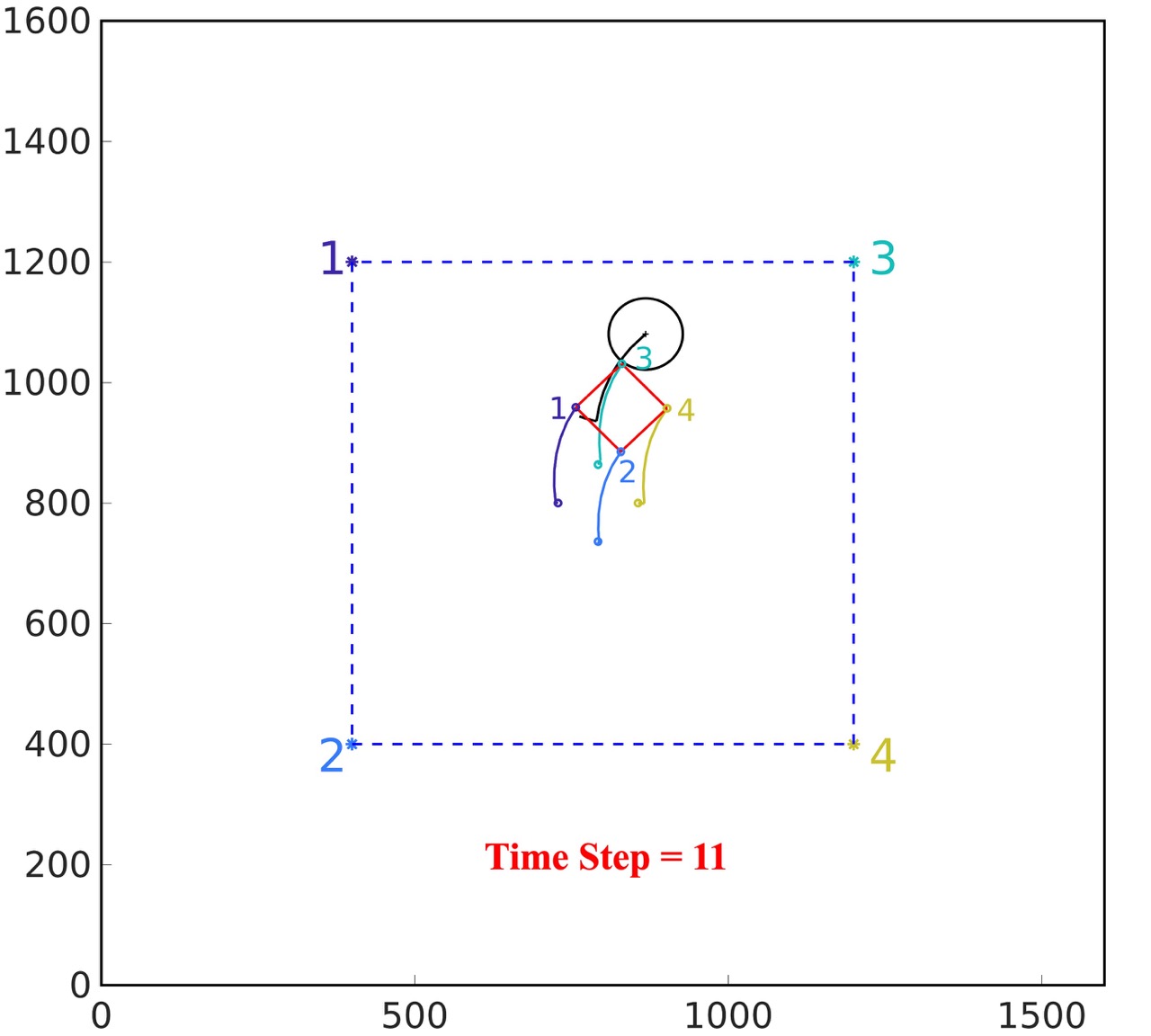}
    \caption{Time step = 11 without noise(CBC)}
    \label{fig2:CBC_11}
  \end{subfigure}
  \begin{subfigure}{0.3\textwidth}
\includegraphics[width=\textwidth]{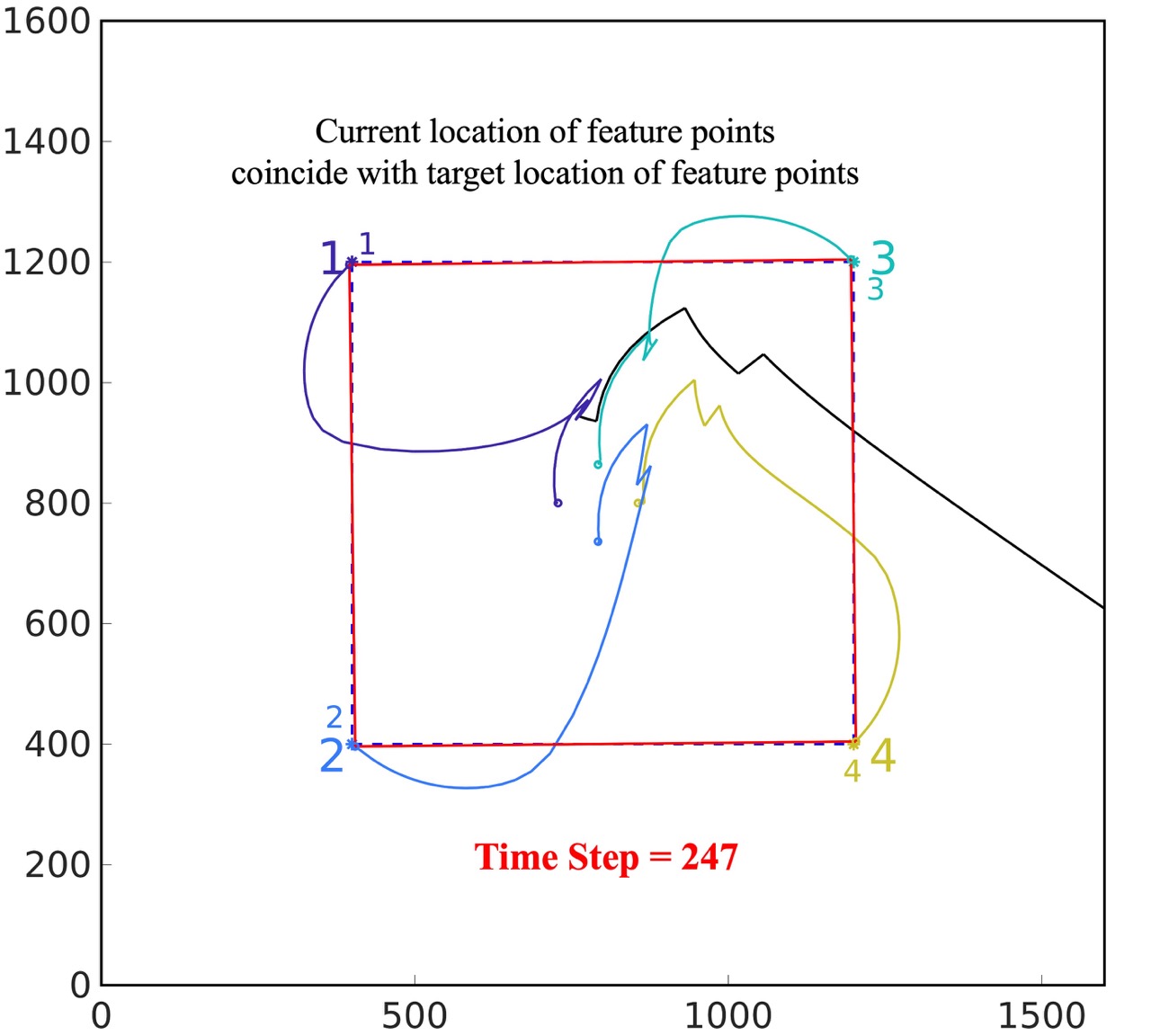}
    \caption{Time step = 247 without noise(CBC)}
    \label{fig2:CBC_end}
  \end{subfigure}
 \captionsetup{skip=0pt}
  \centering
  \begin{subfigure}{0.3\textwidth}
\includegraphics[width=\textwidth]{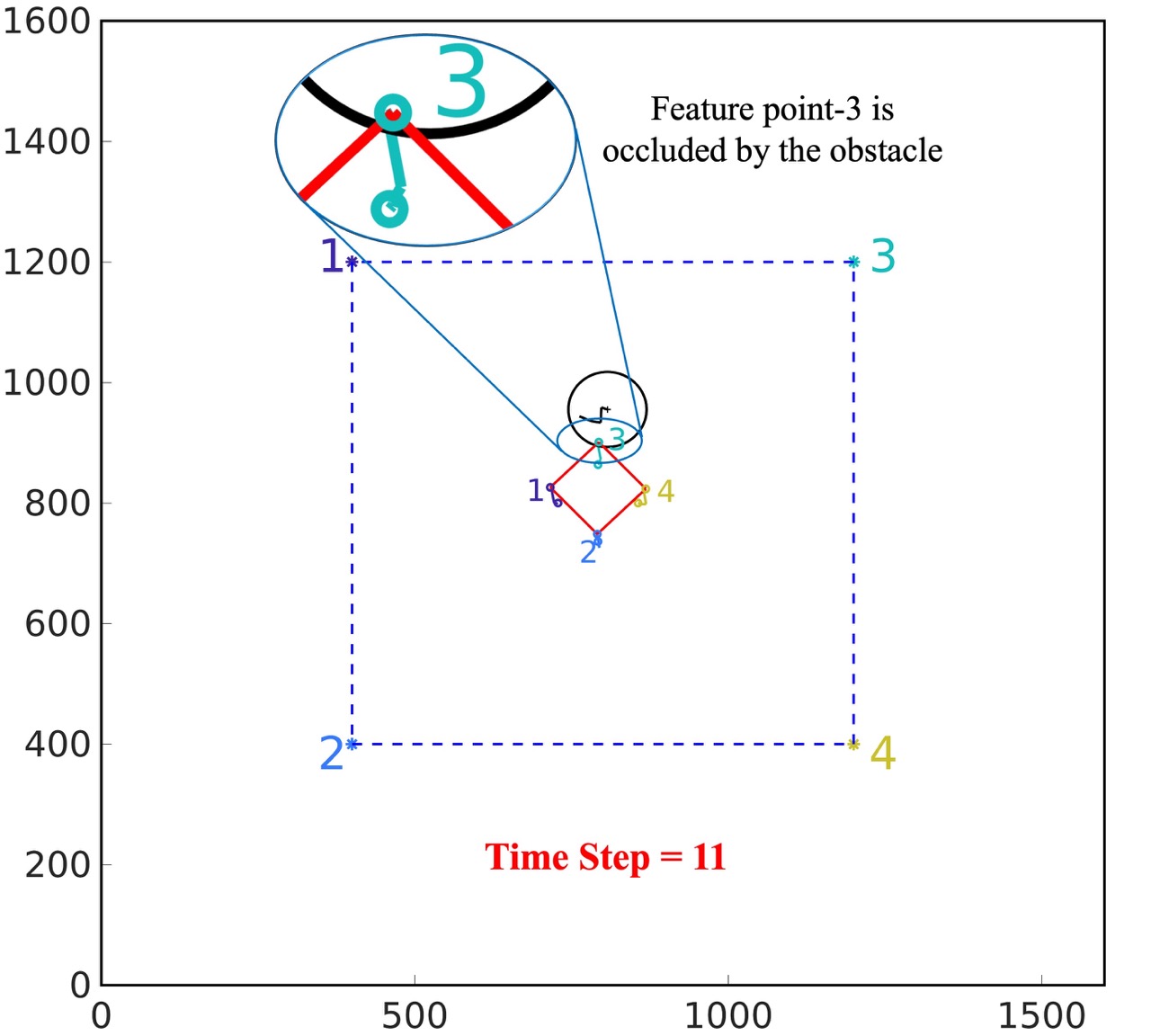}
    \caption{Time step = 11 with noise(CBC)}
    \label{fig2:CBC_noise}
  \end{subfigure}
  \begin{subfigure}{0.3\textwidth}
\includegraphics[width=\textwidth]{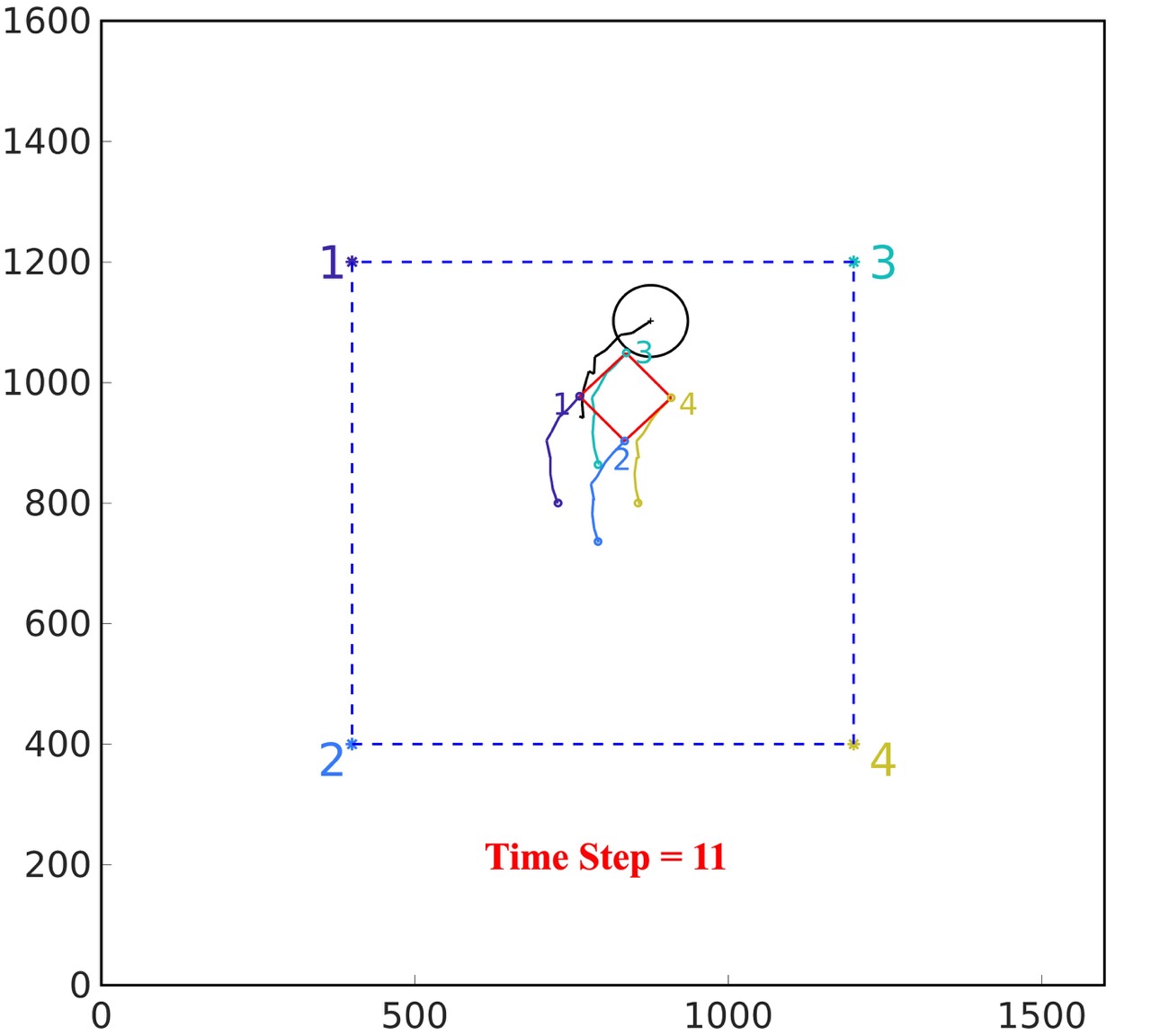}
    \caption{Time step = 11 with noise(PrCBC)}
    \label{fig2:PrCBC_11}
  \end{subfigure}
  \begin{subfigure}{0.3\textwidth}
\includegraphics[width=\textwidth]{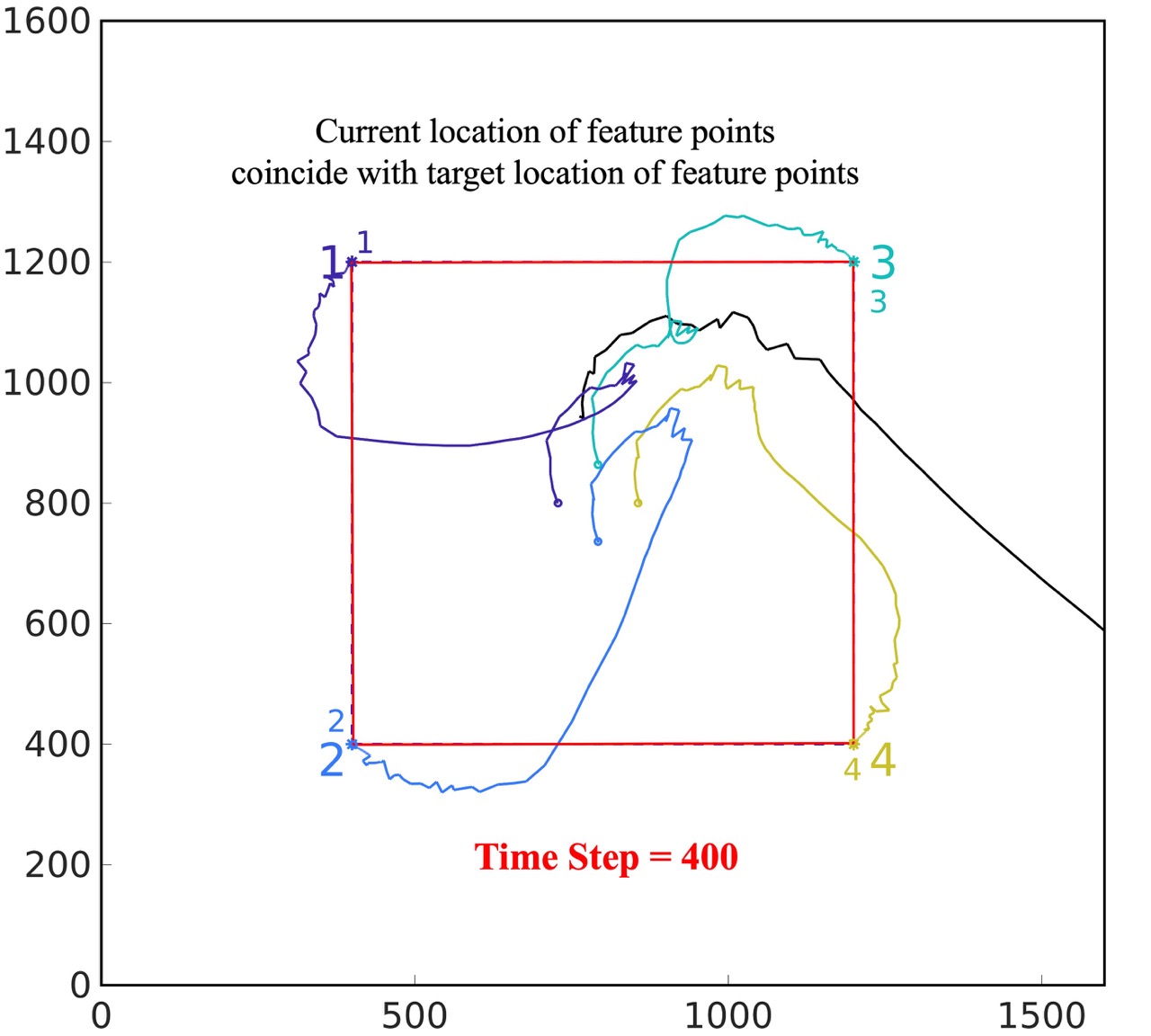}
    \caption{Time step = 400 with noise(PrCBC)}
    \label{fig2:PrCBC_end}
  \end{subfigure}
\caption{The performance of CBC and PrCBC. The obstacle is represented by the black circle. The red and blue squares are the feature points location
of the initial and target location, respectively. The number with different colors are the feature points in the camera. The colored curves indicate the trajectories of corresponding feature points. 
}
  \label{fig:sim_result}
\end{figure*}

\begin{figure*}[!htbp]
\captionsetup{skip=0pt}
  \centering
    \begin{subfigure}{0.23\textwidth}
\includegraphics[width=\textwidth]{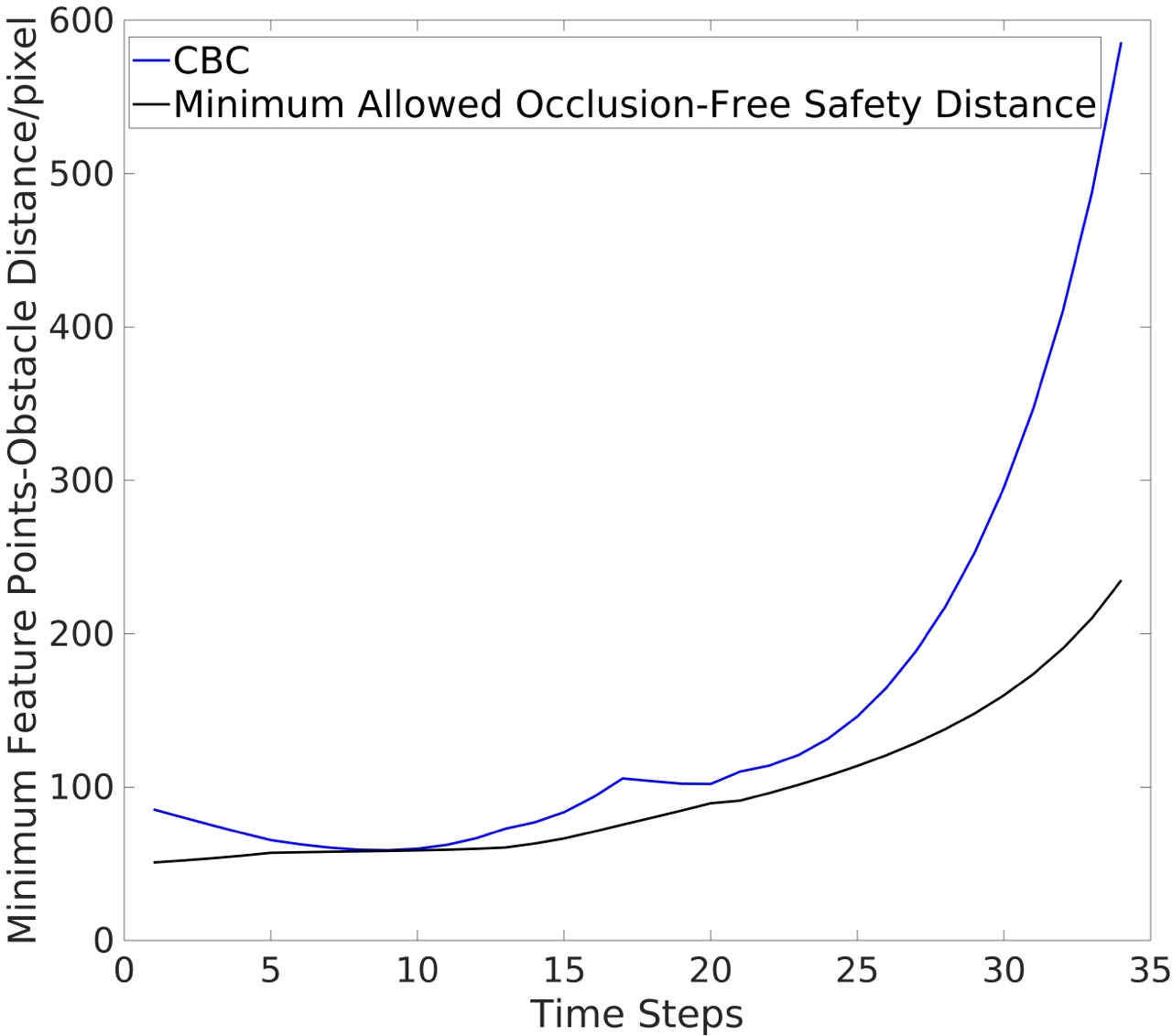}
    \caption{Minimum feature points-obstacle distance without noise (CBC)}
    \label{fig3:CBC}
  \end{subfigure}
  \begin{subfigure}{0.22\textwidth}
\includegraphics[width=\textwidth]{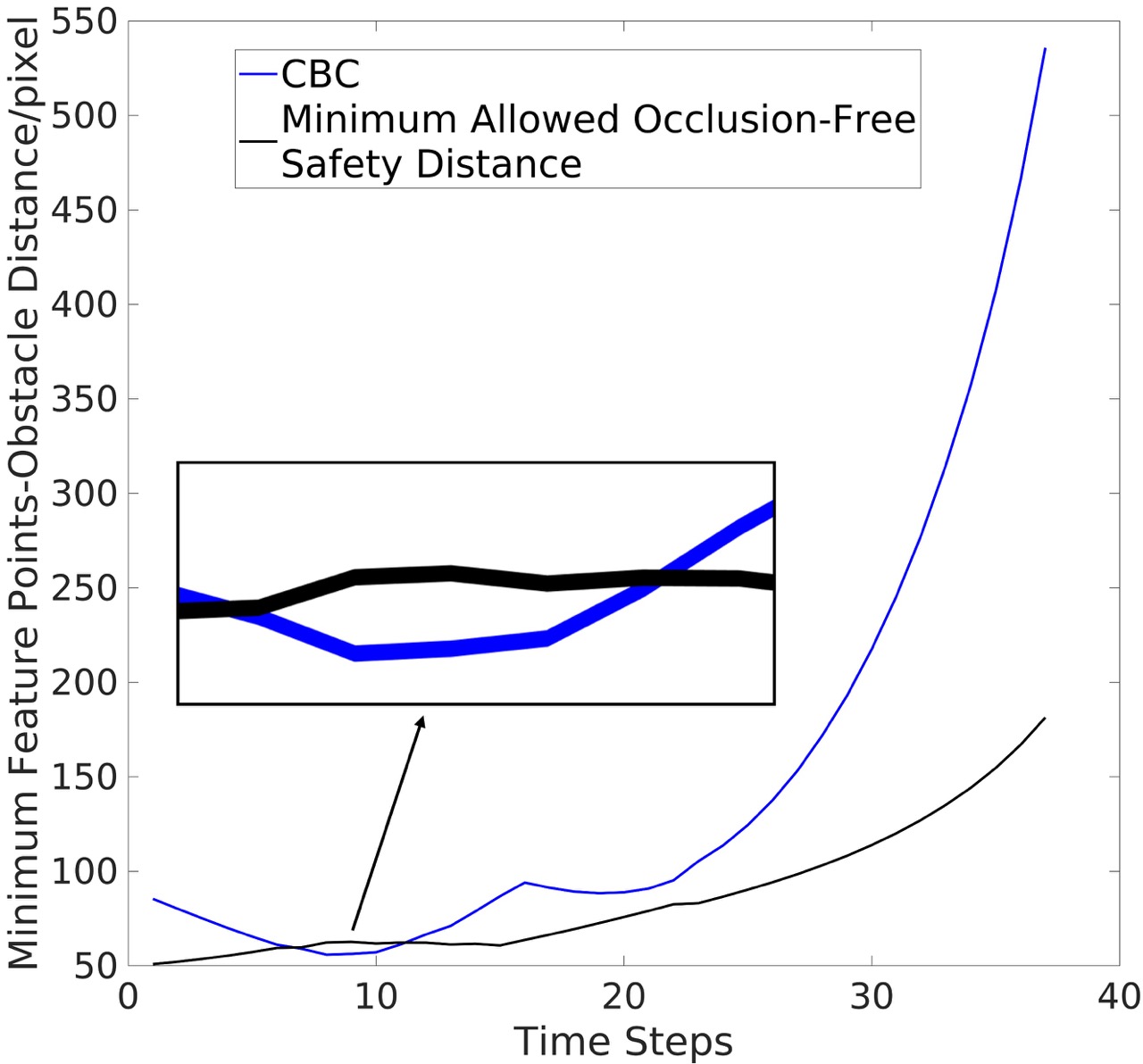}
    \caption{Minimum feature points-obstacle distance with noise (CBC)}
    \label{fig3:CBC_noise}
  \end{subfigure}
  \begin{subfigure}{0.23\textwidth}
\includegraphics[width=\textwidth]{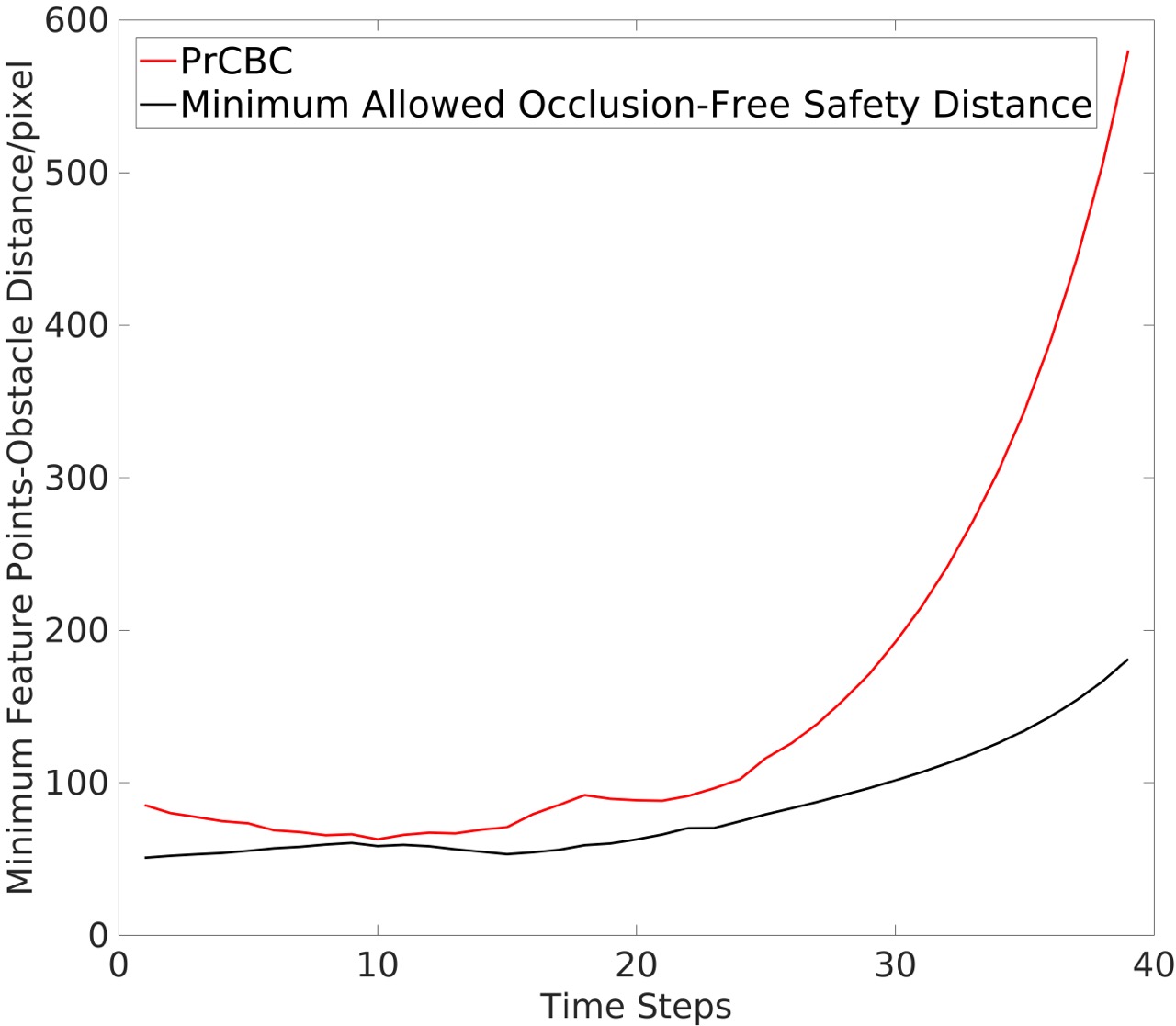}
    \caption{Minimum feature points-obstacle distance with noise (PrCBC)}
    \label{fig3:PrCBC_noise}
  \end{subfigure}
 \begin{subfigure}{0.23\textwidth}
\includegraphics[width=\textwidth]{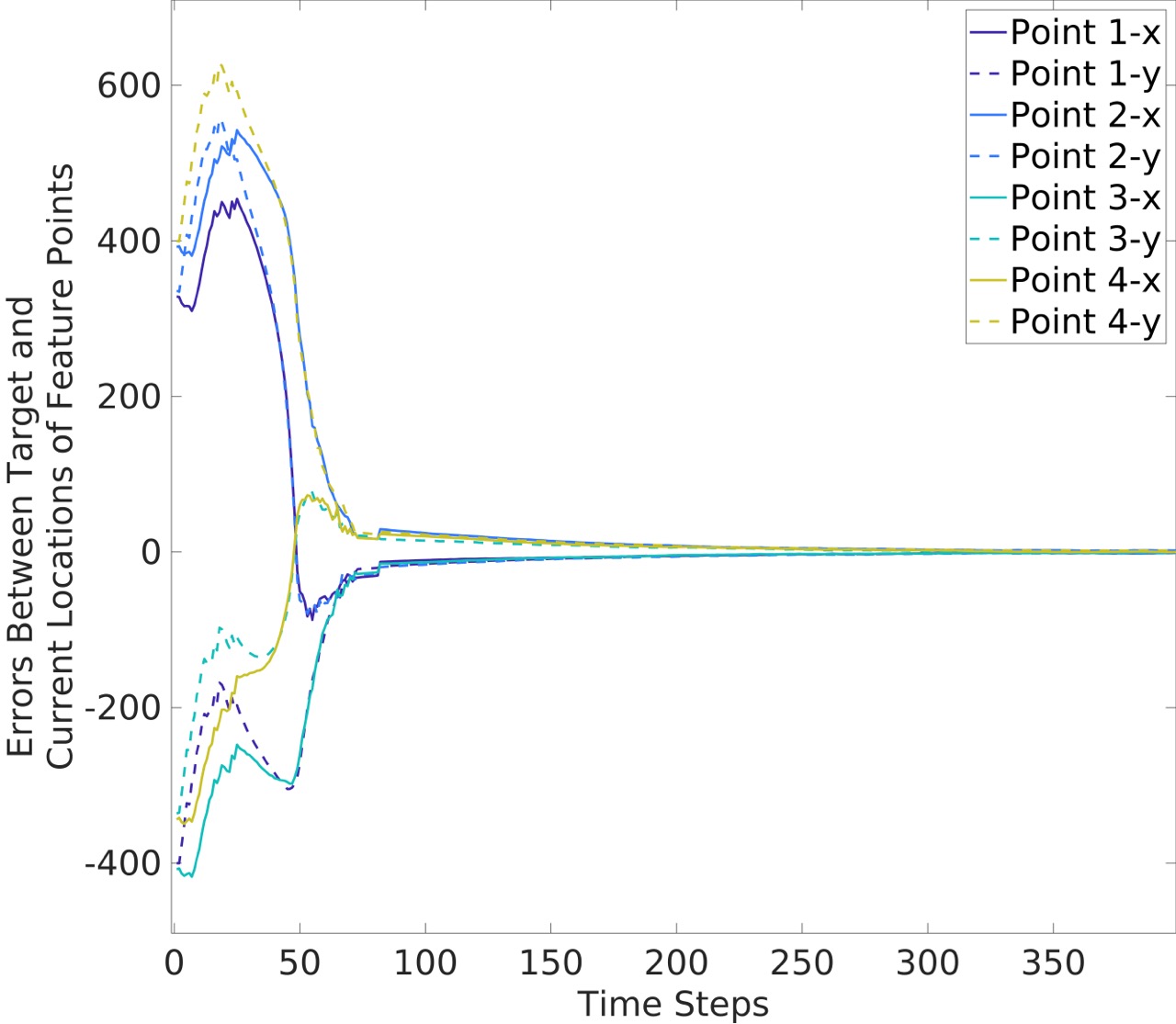}
    \caption{Errors between target and current locations of feature points (PrCBC)}
    \label{fig3:feature_error}
 \end{subfigure}
\caption{Quantitative results of CBC and PrCBC. }
  \label{fig:quantitative}
\end{figure*}

\begin{figure}[!htbp]
\captionsetup{skip=0pt}
  \centering
\includegraphics[width=2.66in]{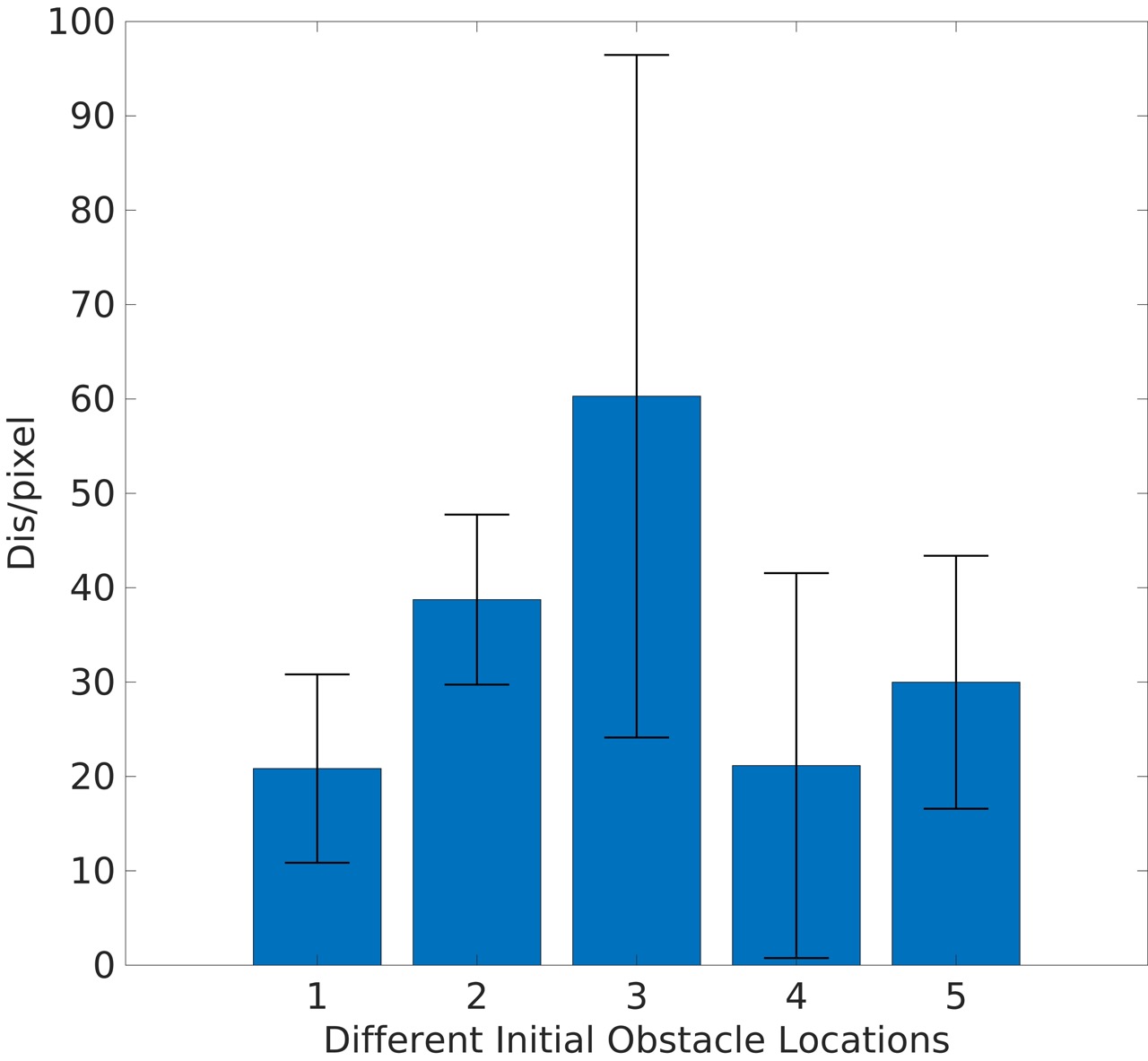}
\caption{Quantitative results summary of PrCBC from 5 different obstacle locations with each having ten random trials. The Y-label $Dis$ is defined as (\ref{Dis}).}
\label{fig4:multi}
\end{figure}

In this section, we use Matlab as the simulation platform and present experimental results to validate the effectiveness of our proposed method. Matlab robotics toolbox \citep{Corke17a} is used to construct the 6-DOF robot simulation model, and the solver IPOPT \citep{wachter2006implementation} is used to generate a high-level planner. 

\subsection{Simulation Performance}
To validate the performance of our experiment, we implement our algorithm under different experiment setups:

\textbf{CBC without noise}: According to the experiment setup shown in Fig.~\ref{fig2:init}, the obstacle is moving in the workspace with initial location $[0.43,0.23,0.10]^{\rm T}$. 
In this experiment, we assume the camera can acquire the accurate coordinates of the feature points and obstacle center. With designed controller CBC, the results are shown in Fig.~\ref{fig2:CBC_11} and Fig.~\ref{fig2:CBC_end}. From Fig.~\ref{fig2:CBC_11} and Fig.~\ref{fig2:CBC_end}, the feature points can avoid occlusion of the obstacle and converge to the pre-defined target locations successfully.

\textbf{CBC with noise}: With the same initial condition of Fig.~\ref{fig2:init}, we assume the pixel coordinates of the feature points and obstacle center are acquired by the camera with Gaussian distributed noise $\mathbf{w}_i,\mathbf{w}_\mathrm{o}\sim N(0,10)$. From Fig.~\ref{fig2:CBC_noise}, it can be observed that the feature point $\#3$ has been occluded by the obstacle in the camera view. 

\textbf{PrCBC with noise}: 
With the same initial condition and measurement noise in Fig.~\ref{fig2:init},
the confidence level is set to be $\sigma = 0.8$.
As shown in Fig.~\ref{fig2:PrCBC_11}, the feature points can avoid the occlusion of the obstacle in the camera view. As shown in Fig,~\ref{fig2:PrCBC_end}, the robot could navigate the camera to the desired location where the four feature points converge to the pre-defined target locations in the camera view.

Hence, the CBC could enforce an occlusion-free IBVS only when accurate feature points information is available. In comparison, the PrCBC is able to guarantee chance-constrained probabilistic occlusion avoidance even in presence of the measurement noise.

\subsection{Quantitative Results}
Next, we present quantitative results from the simulation experiments. As shown in Fig.~\ref{fig3:CBC}, without noise in the pixel coordinates, the CBC method performs well, always ensuring that the minimum feature points-obstacle distance exceeds the occlusion-free safety distance. However, when noise is added, the CBC method is unable to guarantee this, as shown in Fig.~\ref{fig3:CBC_noise} (here we continue running the simulation until the obstacle moving out of the camera view). The PrCBC method, on the other hand, consistently maintains the minimum feature points-obstacle distance above the safety distance, even in the presence of observation noise. 
Fig.~\ref{fig3:feature_error} shows the convergence of the feature points locations in the camera view. It validates that our method can accomplish the IBVS task with occlusion-free performance.

We also performed 50 random trials (5 different initial locations with each one having 10 random trails) under the confidence level of $\sigma=0.9$ to validate the effectiveness of the PrCBC controller in presence of random camera measurement noise. We define $Dis = min(Dis_i(t))$ to express the distance between the feature point and the obstacle edge in the image plane with $Dis_i(t)$ denoted as:
\begin{equation}
\label{Dis}
    Dis_i(t) = \sqrt{{\left \| \mathbf{q}_i(t)-\mathbf{q}_\mathrm{o}(t) \right \|}_2^{2}} - r(t)
\end{equation}
where $r(t) = \frac{fR}{Z_\mathrm{o}(t)} \in \mathbb{R}$ is the obstacle radius in the image plane (pixel scale).
Fig.~\ref{fig4:multi} shows the mean $Dis$ with its corresponding variance,
which verified that 
the feature points would not collide with the obstacle
using the proposed PrCBC, indicating the occlusion-free performance.

\section{Conclusion}
In this paper, we present a control method to address the chance-constrained occlusion avoidance problem between the feature points and obstacle in Image Based Visual Servoing (IBVS) tasks. By adopting the probabilistic control barrier certificates (PrCBC),
we transform the probabilistic occlusion-free conditions to deterministic control constraints, which formally guarantee the performance with satisfying probability under measurement uncertainty. Then we integrate the control constraints with Model Predictive Control (MPC) policy to generate a sequence of optimized control for high-level planning with enforced occlusion avoidance. The simulation results verify the effectiveness of the proposed method. Future works will further explore the real-world applications of the proposed method using feature points such as Scale-Invariant Feature Transform (SIFT), Oriented FAST and Rotated BRIEF (ORB), etc.

\bibliography{IFAC_Camera_Ready}           

\begin{thebibliography}{24}
\providecommand{\natexlab}[1]{#1}
\providecommand{\url}[1]{\texttt{#1}}
\providecommand{\urlprefix}{URL }
\expandafter\ifx\csname urlstyle\endcsname\relax
  \providecommand{\doi}[1]{doi:\discretionary{}{}{}#1}\else
  \providecommand{\doi}{doi:\discretionary{}{}{}\begingroup \urlstyle{rm}\Url}\fi

\bibitem[{Ames et~al.(2019)Ames, Coogan, Egerstedt, Notomista, Sreenath, and Tabuada}]{ames2019control}
Ames, A.D., Coogan, S., Egerstedt, M., Notomista, G., Sreenath, K., and Tabuada, P. (2019).
\newblock Control barrier functions: Theory and applications.
\newblock In \emph{18th European Control Conference (ECC)}, 3420--3431. IEEE.

\bibitem[{Capelli and Sabattini(2020)}]{capelli2020connectivity}
Capelli, B. and Sabattini, L. (2020).
\newblock Connectivity maintenance: Global and optimized approach through control barrier functions.
\newblock In \emph{IEEE International Conference on Robotics and Automation (ICRA)}, 5590--5596. IEEE.

\bibitem[{Chaumette and Hutchinson(2006)}]{chaumette2006visual}
Chaumette, F. and Hutchinson, S. (2006).
\newblock Visual servo control. i. basic approaches.
\newblock \emph{IEEE Robotics \& Automation Magazine}, 13(4), 82--90.

\bibitem[{Chaumette et~al.(2016)Chaumette, Hutchinson, and Corke}]{chaumette2016visual}
Chaumette, F., Hutchinson, S., and Corke, P. (2016).
\newblock Visual servoing.
\newblock In \emph{Springer Handbook of Robotics}, 841--866. Springer.

\bibitem[{Corke(2017)}]{Corke17a}
Corke, P.I. (2017).
\newblock \emph{Robotics, Vision \& Control: Fundamental Algorithms in {MATLAB}}.
\newblock Springer, second edition.
\newblock ISBN 978-3-319-54413-7.

\bibitem[{Corke and Khatib(2011)}]{corke2011robotics}
Corke, P.I. and Khatib, O. (2011).
\newblock \emph{Robotics, vision and control: fundamental algorithms in MATLAB}, volume~73.
\newblock Springer.

\bibitem[{De~Luca et~al.(2008)De~Luca, Oriolo, and Robuffo~Giordano}]{de2008feature}
De~Luca, A., Oriolo, G., and Robuffo~Giordano, P. (2008).
\newblock Feature depth observation for image-based visual servoing: Theory and experiments.
\newblock \emph{The International Journal of Robotics Research}, 27(10), 1093--1116.

\bibitem[{Fleurmond and Cadenat(2016)}]{fleurmond2016handling}
Fleurmond, R. and Cadenat, V. (2016).
\newblock Handling visual features losses during a coordinated vision-based task with a dual-arm robotic system.
\newblock In \emph{2016 European Control Conference (ECC)}, 684--689. IEEE.

\bibitem[{Huang and Mok(2018)}]{huang2018case}
Huang, P.C. and Mok, A.K. (2018).
\newblock A case study of cyber-physical system design: Autonomous pick-and-place robot.
\newblock In \emph{2018 IEEE 24th international conference on embedded and real-time computing systems and applications (RTCSA)}, 22--31. IEEE.

\bibitem[{Kazemi et~al.(2010)Kazemi, Gupta, and Mehrandezh}]{kazemi2010path}
Kazemi, M., Gupta, K., and Mehrandezh, M. (2010).
\newblock Path-planning for visual servoing: A review and issues.
\newblock \emph{Visual Servoing via Advanced Numerical Methods}, 189--207.

\bibitem[{Kermorgant and Chaumette(2013)}]{kermorgant2013dealing}
Kermorgant, O. and Chaumette, F. (2013).
\newblock Dealing with constraints in sensor-based robot control.
\newblock \emph{IEEE Transactions on Robotics}, 30(1), 244--257.

\bibitem[{Landi et~al.(2019)Landi, Ferraguti, Costi, Bonf{\`e}, and Secchi}]{landi2019safety}
Landi, C.T., Ferraguti, F., Costi, S., Bonf{\`e}, M., and Secchi, C. (2019).
\newblock Safety barrier functions for human-robot interaction with industrial manipulators.
\newblock In \emph{2019 18th European Control Conference (ECC)}, 2565--2570. IEEE.

\bibitem[{Li et~al.(2020)Li, Chiu, and Li}]{li2020accelerated}
Li, W., Chiu, P.W.Y., and Li, Z. (2020).
\newblock An accelerated finite-time convergent neural network for visual servoing of a flexible surgical endoscope with physical and rcm constraints.
\newblock \emph{IEEE transactions on neural networks and learning systems}, 31(12), 5272--5284.

\bibitem[{Luo et~al.(2020{\natexlab{a}})Luo, Sun, and Kapoor}]{luo2020multi}
Luo, W., Sun, W., and Kapoor, A. (2020{\natexlab{a}}).
\newblock Multi-robot collision avoidance under uncertainty with probabilistic safety barrier certificates.
\newblock \emph{Advances in Neural Information Processing Systems}, 33, 372--383.

\bibitem[{Luo et~al.(2020{\natexlab{b}})Luo, Yi, and Sycara}]{luo2020behavior}
Luo, W., Yi, S., and Sycara, K. (2020{\natexlab{b}}).
\newblock Behavior mixing with minimum global and subgroup connectivity maintenance for large-scale multi-robot systems.
\newblock In \emph{2020 IEEE International Conference on Robotics and Automation (ICRA)}, 9845--9851. IEEE.

\bibitem[{Lyu et~al.(2021)Lyu, Luo, and Dolan}]{lyu2021probabilistic}
Lyu, Y., Luo, W., and Dolan, J.M. (2021).
\newblock Probabilistic safety-assured adaptive merging control for autonomous vehicles.
\newblock In \emph{2021 IEEE International Conference on Robotics and Automation (ICRA)}, 10764--10770. IEEE.

\bibitem[{Marchand et~al.(2005)Marchand, Spindler, and Chaumette}]{Marchand05b}
Marchand, E., Spindler, F., and Chaumette, F. (2005).
\newblock Visp for visual servoing: a generic software platform with a wide class of robot control skills.
\newblock \emph{IEEE Robotics and Automation Magazine}, 12(4), 40--52.

\bibitem[{Mezouar and Chaumette(2002)}]{mezouar2002avoiding}
Mezouar, Y. and Chaumette, F. (2002).
\newblock Avoiding self-occlusions and preserving visibility by path planning in the image.
\newblock \emph{Robotics and Autonomous Systems}, 41(2-3), 77--87.

\bibitem[{Nicolis et~al.(2018)Nicolis, Palumbo, Zanchettin, and Rocco}]{nicolis2018occlusion}
Nicolis, D., Palumbo, M., Zanchettin, A.M., and Rocco, P. (2018).
\newblock Occlusion-free visual servoing for the shared autonomy teleoperation of dual-arm robots.
\newblock \emph{IEEE Robotics and Automation Letters}, 3(2), 796--803.

\bibitem[{Saragih et~al.(2019)Saragih, Kinasih, Machbub, Rusmin, and Rohman}]{saragih2019visual}
Saragih, C.F.D., Kinasih, F.M.T.R., Machbub, C., Rusmin, P.H., and Rohman, A.S. (2019).
\newblock Visual servo application using model predictive control (mpc) method on pan-tilt camera platform.
\newblock In \emph{2019 6th International Conference on Instrumentation, Control, and Automation (ICA)}, 1--7. IEEE.

\bibitem[{W{\"a}chter and Biegler(2006)}]{wachter2006implementation}
W{\"a}chter, A. and Biegler, L.T. (2006).
\newblock On the implementation of an interior-point filter line-search algorithm for large-scale nonlinear programming.
\newblock \emph{Mathematical programming}, 106(1), 25--57.

\bibitem[{Xiao et~al.(2022)Xiao, Belta, and Cassandras}]{xiao2022sufficient}
Xiao, W., Belta, C.A., and Cassandras, C.G. (2022).
\newblock Sufficient conditions for feasibility of optimal control problems using control barrier functions.
\newblock \emph{Automatica}, 135, 109960.

\bibitem[{Xu et~al.(2017)Xu, Waters, Pickem, Glotfelter, Egerstedt, Tabuada, Grizzle, and Ames}]{xu2017realizing}
Xu, X., Waters, T., Pickem, D., Glotfelter, P., Egerstedt, M., Tabuada, P., Grizzle, J.W., and Ames, A.D. (2017).
\newblock Realizing simultaneous lane keeping and adaptive speed regulation on accessible mobile robot testbeds.
\newblock In \emph{2017 IEEE Conference on Control Technology and Applications (CCTA)}, 1769--1775. IEEE.

\bibitem[{Zheng et~al.(2019)Zheng, Wang, Wang, Zhang, and Chen}]{zheng2019toward}
Zheng, D., Wang, H., Wang, J., Zhang, X., and Chen, W. (2019).
\newblock Toward visibility guaranteed visual servoing control of quadrotor uavs.
\newblock \emph{IEEE/ASME Transactions on Mechatronics}, 24(3), 1087--1095.

\end{thebibliography}
\end{document}